\documentclass{article}
\usepackage{ijcai24}

\usepackage{times}
\usepackage{soul}
\usepackage{url}
\usepackage[utf8]{inputenc}
\usepackage[small]{caption}
\usepackage{graphicx}
\usepackage{amsmath}
\usepackage{amsthm}
\usepackage{booktabs}
\usepackage{algorithm}
\usepackage{algorithmic}
\usepackage[switch]{lineno}
\usepackage{microtype}
\usepackage[usenames,svgnames,table]{xcolor}
\usepackage{makecell}
\usepackage[hidelinks]{hyperref}
\usepackage{mydef}
\usepackage{multirow}
\usepackage{xcolor}
\usepackage{amssymb}
\usepackage{tikz}
\usepackage[edges]{forest}
\usepackage{array}
\usepackage{booktabs}
\usepackage{caption}
\usepackage{cellspace}   % Ensures a minimal spacing of table cells
\definecolor{tablebordercolor}{rgb}{0.373, 0.620, 0.627} % Light blue color
\title{AI-Enhanced Virtual Reality in Medicine: A Comprehensive Survey}

\author{
Yixuan Wu$^{1,2,3}$
\and
Kaiyuan Hu$^{4}$\and
Danny Z. Chen$^5$\And
Jian Wu$^{1,2}$\\
\affiliations
$^1$School of Public Health, Zhejiang University, Hangzhou 310058, China\\
$^2$State Key Laboratory of Transvascular Implantation Devices of The Second Affiliated Hospital, Zhejiang University School of Medicine, Hangzhou 310009, China\\
$^3$Institute of Wenzhou, Zhejiang University, Hangzhou 325036, China\\
$^4$School of Science and Engineering, The Chinese University of Hong Kong, Shenzhen, China\\
$^5$Department of Computer Science and Engineering, University of Notre Dame, USA\\
\emails
wyx\_chloe@zju.edu.cn,
kaiyuanhu@link.cuhk.edu.cn,
dchen@nd.edu,
wujian2000@zju.edu.cn\\
}

\begin{document}

\maketitle
% \renewcommand{\thefootnote}{\fnsymbol{footnote}}  
% \footnotetext[1]{These authors contributed equally to this work.}  
% \footnotetext[2]{Corresponding authors.}  

\begin{abstract}
    With the rapid advance of computer graphics and artificial intelligence technologies, the ways we interact with the world have undergone a transformative shift. Virtual Reality (VR) technology, aided by artificial intelligence (AI), has emerged as a dominant interaction media in multiple application areas, thanks to its advantage of providing users with immersive experiences. Among those applications, medicine is considered one of the most promising areas. 
    In this paper, we present a comprehensive examination of the burgeoning field of AI-enhanced VR applications in medical care and services. 
    By introducing a systematic taxonomy, we meticulously classify the pertinent techniques and applications into three well-defined categories based on different phases of medical diagnosis and treatment: Visualization Enhancement, VR-related Medical Data Processing, and VR-assisted Intervention. This categorization enables a structured exploration of the diverse roles that AI-powered VR plays in the medical domain, providing a framework for a more comprehensive understanding and evaluation of these technologies.
    To our best knowledge, this work is the first systematic survey of AI-powered VR systems in medical settings, laying a foundation for future research in this interdisciplinary domain.
    
\end{abstract}

\begin{figure*}
    \centering
    \includegraphics[width=1\textwidth]{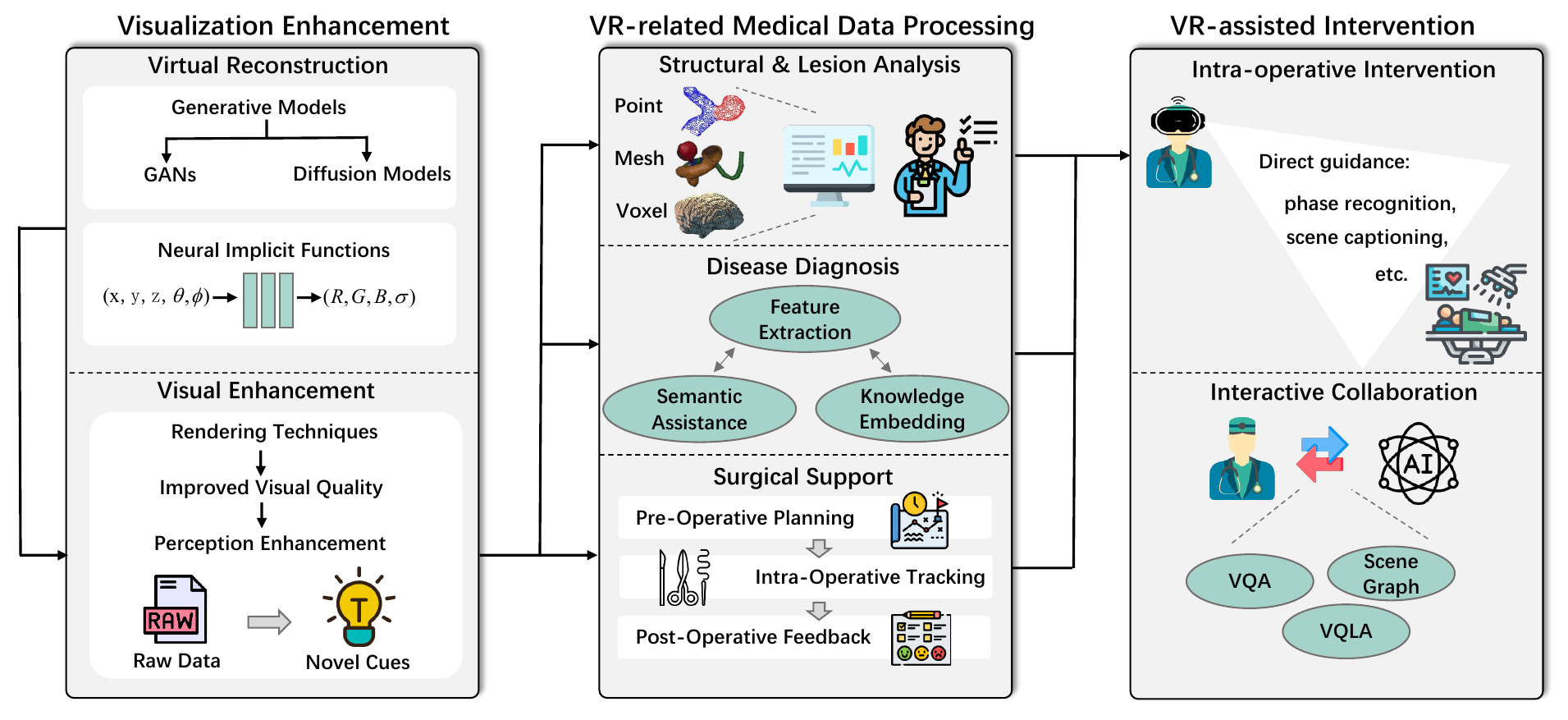}
    \caption{Demonstrating the workflow of AI-enhanced VR-assisted medical services. First, Visualization Enhancement focuses on the augmentation of a user’s visual and sensory perception within a virtual space. Next, VR-related Medical Data Processing discusses how VR, combined with AI, facilitates in-depth structural and lesion analysis, enhances disease diagnosis, and supports the entire stages of surgical procedures. Finally, VR-assisted Intervention covers the utilization of VR for direct procedural guidance, intra-operative intervention, and fostering interactive collaborations among medical professionals.}
    \label{fig:framework}
\end{figure*}

\section{Introduction}

The intersection of artificial intelligence (AI) and virtual reality (VR) is an emerging frontier in the realm of medical technologies. Recent advances in these fields have given rise to innovative applications that promise to reshape various aspects of medical care. The integration of AI and VR brings forth a unique confluence of data-driven analytics and immersive experiences, making it a critical area of study and application in contemporary healthcare.
AI’s contribution to VR in medicine ranges from enhancing diagnostic precision to offering new paradigms in patient care. These technologies are not only redefining existing medical procedures but also paving the way for novel treatment methods. 
The purpose of this survey is to delve into the technical nuances, practice workflows, and diverse application scenarios of AI-powered VR in medical settings, examining their impacts on the efficiency, accuracy, and effectiveness of healthcare services.

To systematically understand and analyze the role of AI in medical VR, this paper introduces a taxonomy categorizing the applications into three primary categories: Visualization Enhancement, VR-related Medical Data Processing, and VR-assisted Intervention (see Figure~\ref{fig:framework}). 
This classification allows for a comprehensive overview of the current state of the art of AI in VR applications tailored to medical needs, highlighting each category’s unique contributions to healthcare. 
By categorizing these applications based on their functions and utilities in the spectrum of patient care, we aim to provide an 
extensive examination that facilitates easier comprehension and identification of existing research gaps.

Visualization Enhancement focuses on the augmentation of a user's visual and sensory perception within a virtual space. This enhancement is crucial in complex medical procedures where understanding intricate anatomical structures and spatial relationships is imperative. We delve into the subcategories of virtual object reconstruction and visual enhancement, both pivotal in improving clinical outcomes and professional training.
%Furthermore, 
VR-related Medical Data Processing addresses the advanced capabilities of VR systems to analyze and interpret complex medical data. This category discusses how VR, combined with AI, facilitates in-depth structural and lesion analysis, enhances disease diagnosis, and supports the entire stages of surgical procedures. It underscores the transformation from traditional 2D data interpretation to a more dynamic, 3D analytical approach.
%Finally, 
VR-assisted Intervention demonstrates the practical application of AI-driven VR in real-time and interactive medical scenarios. This category covers the utilization of VR for direct procedural guidance, intra-operative intervention, and fostering interactive collaborations among medical professionals. It exemplifies the potential of VR to not only assist but also augment medical practice in live environments.

To our best knowledge, this is the first systematic survey of AI-powered VR technologies in medical settings, laying a foundation for future research in this interdisciplinary domain. The scope of this survey extends beyond mere categorization of current techniques and applications. It also explores the potential future trajectories of AI in VR for medicine, contemplating how ongoing advancements might unfold. The discussion encompasses not only technological aspects but also considers ethical, legal, and practical implications of deploying such advanced systems in sensitive medical environments.

% %How AI tackles challenges in VR applications
% Virtual Reality (VR) promises the capability of providing users with an immersive and interactive experience, impacting not only entertainment and education but also extending its potential to revolutionize more industries like healthcare. However, the integration of such features poses multiple challenges regarding high computation consumption, large network bandwidth cost, and low latency requirements. To address these hurdles, recent advancements in VR technology have employed various machine-learning methods such as Recurrent Neural Networks (RNN) for precise pose prediction, Deep-Reinforcement-Learning (DRL) for network adaptation to optimize data transmission, or Convolutional Neural Networks (CNN) for fast image processing, ensuring instant scene understanding or object detection. These machine-learning techniques play a pivotal role in pushing the boundaries of VR capabilities and making strides toward more efficient, adaptive, and seamless virtual experiences.

\tikzstyle{leaf}=[draw=hiddendraw,
    rounded corners,minimum height=1em,
    fill=hidden-orange!40,text opacity=1, align=center,
    fill opacity=.5,  text=black,align=left,font=\scriptsize,
    inner xsep=3pt,
    inner ysep=2.5pt,
    ]
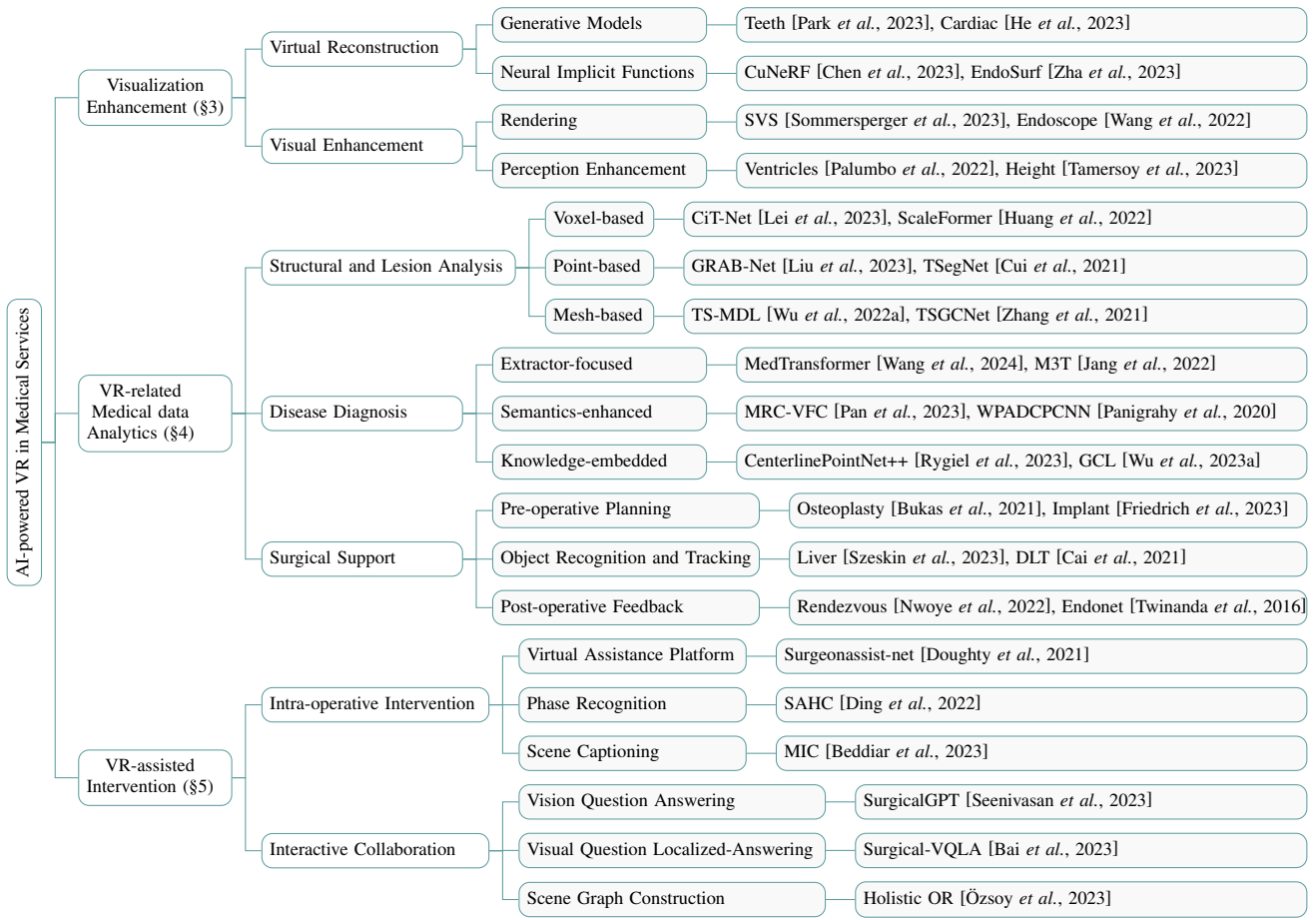
\begin{figure*}[ht]
\centering
\begin{forest}
  for tree={
  forked edges,
  grow=east,
  reversed=true,
  anchor=base west,
  parent anchor=east,
  child anchor=west,
  base=middle,
  font=\scriptsize,
  rectangle,
  draw=hiddendraw,
  rounded corners,align=left,
  minimum width=2em,
    s sep=5pt,
    inner xsep=3pt,
    inner ysep=2.5pt,
  },
  where level=1{text width=4.5em}{},
  where level=2{text width=6em,font=\scriptsize}{},
  where level=3{font=\scriptsize}{},
  where level=4{font=\scriptsize}{},
  where level=5{font=\scriptsize}{},
  [AI-powered VR in Medical Services,rotate=90,anchor=north,edge=hiddendraw
    [Visualization \\
    Enhancement (\S\ref{part1}),edge=hiddendraw,align=center,text width=5.2em
        [Virtual Reconstruction, text width=7em, edge=hiddendraw
            [Generative Models,leaf,text width=7.5em, edge=hiddendraw
                        [Teeth~\cite{park20233d}{,} Cardiac~\cite{he2023dmcvr},leaf,text width=21em, edge=hiddendraw]
                        ]
            [Neural Implicit Functions,leaf,text width=7.5em, edge=hiddendraw
                        [CuNeRF~\cite{chen2023cunerf}{,} EndoSurf~\cite{zha2023endosurf},leaf,text width=21em, edge=hiddendraw]
                        ]
        ]
        [Visual Enhancement, text width=7em, edge=hiddendraw
            [Rendering,leaf,text width=7.5em, edge=hiddendraw
                                    [SVS~\cite{sommersperger2023semantic}{,} Endoscope~\cite{wang2022neural},leaf,text width=21em, edge=hiddendraw]
            ]
            [Perception Enhancement,leaf,text width=7.5em, edge=hiddendraw
                                    [Ventricles~\cite{palumbo2022mixed}{,} Height~\cite{tamersoy2023accurate},leaf,text width=21em, edge=hiddendraw]
                                    ]
        ]
    ]
    [VR-related\\
    Medical data \\
    Analytics (\S\ref{part2}),edge=hiddendraw,align=center,text width=5.2em
     [Structural and Lesion Analysis, text width=9em, edge=hiddendraw
            [Voxel-based,leaf,text width=3.5em, edge=hiddendraw
            [CiT-Net~\cite{lei2023cit}{,} ScaleFormer~\cite{huang2022scaleformer},leaf,text width=23em, edge=hiddendraw]
            ]
            [Point-based,leaf,text width=3.5em, edge=hiddendraw
            [GRAB-Net~\cite{liu2023grab}{,} TSegNet~\cite{cui2021tsegnet} ,leaf,text width=23em, edge=hiddendraw]
            ]
            [Mesh-based,leaf,text width=3.5em, edge=hiddendraw
            [TS-MDL~\cite{wu2022two}{,} TSGCNet~\cite{zhang2021tsgcnet},leaf,text width=23em, edge=hiddendraw]
            ]
      ]
    [Disease Diagnosis, text width=7em, edge=hiddendraw
            [Extractor-focused,leaf,text width=7.5em, edge=hiddendraw
                [MedTransformer~\cite{wang2024medtransformer}{,} M3T~\cite{jang2022m3t} ,leaf,text width=21em, edge=hiddendraw]
                        ]
            [Semantics-enhanced,leaf,text width=7.5em, edge=hiddendraw
                [MRC-VFC~\cite{pan2023combat}{,} WPADCPCNN~\cite{panigrahy2020mri} ,leaf,text width=21em, edge=hiddendraw]
                        ]
            [Knowledge-embedded,leaf,text width=7.5em, edge=hiddendraw
                [CenterlinePointNet++~\cite{rygiel2023centerlinepointnet}{,} GCL~\cite{wu2023gcl} ,leaf,text width=21em, edge=hiddendraw]
                        ]
     ]
    [Surgical Support, text width=7em, edge=hiddendraw
            [Pre-operative Planning,leaf,text width=9.5em, edge=hiddendraw
                            [Osteoplasty~\cite{bukas2021patient}{,} Implant~\cite{friedrich2023point} ,leaf,text width=19em, edge=hiddendraw]]
            [Object Recognition and Tracking,leaf,text width=9.5em, edge=hiddendraw
                            [Liver~\cite{szeskin2023liver}{,} DLT~\cite{cai2021deep} ,leaf,text width=19em, edge=hiddendraw]]
            [Post-operative Feedback,leaf,text width=9.5em, edge=hiddendraw
                            [Rendezvous~\cite{nwoye2022rendezvous}{,} Endonet~\cite{twinanda2016endonet} ,leaf,text width=19em, edge=hiddendraw]]
     ]
    ]
    [VR-assisted \\
    Intervention (\S\ref{part3}), edge=hiddendraw,align=center,text width=5.2em
      [Intra-operative  
      Intervention,text width=8em, edge=hiddendraw
            [Virtual Assistance Platform,leaf,text width=8em, edge=hiddendraw
            [Surgeonassist-net~\cite{doughty2021surgeonassist},leaf,text width=19.5em, edge=hiddendraw]
            ]
            [Phase Recognition,leaf,text width=8em, edge=hiddendraw
            [SAHC~\cite{ding2022exploring},leaf,text width=19.5em, edge=hiddendraw]
            ]
            [Scene Captioning,leaf,text width=8em, edge=hiddendraw
            [MIC~\cite{beddiar2023automatic},leaf,text width=19.5em, edge=hiddendraw]
            ]
      ]
      [Interactive 
      Collaboration,text width=8em, edge=hiddendraw
            [Vision Question Answering,leaf,text width=11em, edge=hiddendraw
            [SurgicalGPT~\cite{seenivasan2023surgicalgpt},leaf,text width=16.5em, edge=hiddendraw]
            ]
            [Visual Question Localized-Answering,leaf,text width=11em, edge=hiddendraw
            [Surgical-VQLA~\cite{bai2023surgical},leaf,text width=16.5em, edge=hiddendraw]
            ]
            [Scene Graph Construction,leaf,text width=11em, edge=hiddendraw
            [Holistic OR~\cite{ozsoy2023holistic},leaf,text width=16.5em, edge=hiddendraw]
            ]
      ]
    ]
    ]
\end{forest}
\caption{A taxonomy of AI-powered VR in medical services with representative examples.}
\label{taxonomy}
\end{figure*}

\section{Taxonomy}
The objective of the taxonomy is to group AI-powered VR applications in medical services with similar goals into the same category, facilitating in-depth investigation in subsequent studies. We classify the techniques and applications into three distinct categories based on diagnosis and treatment procedures: \textit{Visualization Enhancement}, \textit{VR-related Medical Data Processing}, and \textit{VR-assisted Intervention}. A visual presentation of the taxonomy is shown in Figure~\ref{taxonomy}.

\subsection{Visualization Enhancement}
The rapid advance of AI technologies offers VR platforms the capability of enhancing medical professionals' visual perception during treatment and surgical stages. The techniques of visualization enhancement can be divided into the following two subcategories. 

\noindent\textbf{Virtual Reconstruction} enables medical professionals to visualize medical data in a more intuitive manner, enhancing their understanding and interpretation of complex anatomical structures. By reconstructing virtual objects (e.g., organs or anatomical models) in a VR environment, professionals can gain a clearer and more comprehensive view of the patient's condition, aiding treatment planning and decision-making.

\noindent\textbf{Visual Enhancement} focuses on improving the visual perception of professionals during the training or surgical phase. By immersing the professionals in augmented or virtual surgical scenes, such techniques create an immersive visual perception that enables them to perform treatment procedures with greater precision, resulting in improved treatment outcomes.

\subsection{VR-related Medical Data Processing}
Enhanced vision in a virtual environment offers an additional analytical capability for processing medical data. By leveraging the immersive and interactive aspect of virtual reality, professionals gain access to a wealth of additional visual information that surpasses the limitations of traditional 2D data usage in different treatment phases, as follows.

\noindent\textbf{Structural and Lesion Analysis.}
In VR medical contexts, the utilized data formats such as point cloud, mesh, and voxel facilitate AI-powered systems to conduct more comprehensive analysis of anatomical structures and lesion situations, thereby providing additional cognitive information for accurate diagnosis.

\noindent\textbf{Disease Diagnosis.}
Comprehensive analysis of VR-based medical data serves as an underlying foundation for disease diagnosis through multiple methods, including semantic segmentation, feature extraction, and knowledge embedding.

\noindent\textbf{Surgical Support.}
Building upon data analysis and diagnosis, surgical support in VR enhances the precision and effectiveness of surgical procedures. Current progress has covered multiple aspects including pre-operative planning, intra-operative recognition and tracking, and post-operative feedback.

\subsection{VR-assisted Intervention}
VR-assisted intervention harnesses the power of AI-enhanced visualization and analysis technologies, showcasing its immense potential in augmenting professionals' capabilities during the inspection and surgical phases. This topic explores how VR technologies directly provide guidance in intra-operative interventions and deliver context-aware feedback in an interactive and collaborative manner.

\noindent\textbf{Intra-operative Intervention} offers professionals direct guidance and assistance by integrating AI-powered functionalities into VR platforms such as object segmentation, phase recognition, etc. This encompasses the integration of diverse input data formats and platforms.

\noindent\textbf{Interactive Collaboration} provides interactive feedback to professionals by leveraging advancements in human-machine collaboration. Techniques such as Vision Question Answering (VQA) and Visual Question Localized-Answering (VQLA) have played a key role in this progress.

\section{Visualization Enhancement}
\label{part1}
%\section{Scene Enhancement}
Enhanced visualization capability is one of the key advantages of VR technologies, which also shows great potential in the enhancement of medical services. 
In this section, we cover two key aspects in the application of visualization enhancement: \textit{virtual reconstruction} and \textit{visual enhancement}.

\subsection{Virtual Reconstruction}
% Describe current reconstruction methods in VR medical services
VR technologies offer the capability to provide users with an immersive experience with infinite viewpoints, enabling them to visualize an object freely. This feature shows great potential in medical services, which enables professionals to visualize complex medical data in a more intuitive manner. Traditional medical data acquisition methods like X-ray and ultrasound mostly provide only flattened 2D images which limit the effectiveness of diagnosis. In contrast, AI-powered reconstruction methods provide solutions for reconstructing complete objects with high quality, which can be visualized with VR devices to provide professionals with a better perception of the target objects. 
Compared with traditional 2D monitoring methods, 3D reconstruction offers notable advantages since it allows users to observe a surgical site from any viewpoint, which dramatically benefits downstream medical applications such as surgeon-centered augmented reality~\cite{nicolau2011augmented} and virtual reality~\cite{chong2022virtual}. 
Multiple learning-based reconstruction methodologies have become prominent, with \textit{neural implicit functions} and \textit{generative models} as the two main approaches.

\noindent\textbf{Generative Models.}
In practical medical inspection scenarios, due to the limitations of safe radiation exposure, visual occlusion, or anisotropic spatial resolution, the reconstruction of high-quality 3D models in medical scenarios has been challenging. 
Thanks to the capability to synthesize image details with higher fidelity, generative adversarial networks (GANs)~\cite{GAN} have been widely adopted to reconstruct high-quality medical images~\cite{kaplan2019full}. 
In \cite{lei2019whole}, a cycle-consistent generative adversarial network (Cycle GAN) model was proposed to estimate diagnostic quality PET images using low count data. 
Recently, due to improved sample quality and higher log-likelihood scores compared to GANs, diffusion probabilistic models~\cite{ho2020denoising} have emerged as a compelling alternative. 
Such methods have been employed in 3D teeth reconstruction~\cite{park20233d} and cardiac volume reconstruction~\cite{he2023dmcvr}. Likewise, to tackle the problem of inadequate natural medical data for temporal dynamics analysis and disease progression monitoring, diffusion models have been exploited to generate 4D volumetric data. In~\cite{kim2022diffusion}, a diffusion deformable model (DDM) was proposed to generate intermediate temporal volumes between source and target volumes.

\begin{table*}[ht]
\centering
\captionsetup{justification=centering}
\arrayrulecolor{tablebordercolor}
\begin{tabular}{ | Sc | Sc | Sc | Sc | Sc | m{2cm} | }
\hline
\rowcolor{gray!25}
\textbf{Representation} & \textbf{Size} & \textbf{Visual Quality} & \textbf{Computing Resource} & \textbf{Editability} & \textbf{Visualization} \\
% \textbf{Representation} & \textbf{Size} & \textbf{Visual} \\ \textbf{Quality} & \textbf{Computing} \\ \textbf{Resource} & \textbf{Editability} & \textbf{Visualization} \\
\hline
Point Cloud~\cite{yang2020intra} & Large & Low & Low & Easy & \includegraphics[width=2cm,height=2cm,keepaspectratio]{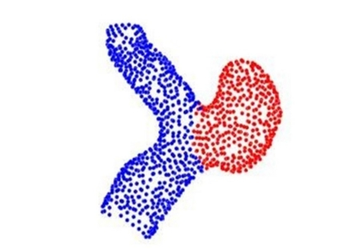} \\
\hline
Mesh~\cite{wu2022two} & Medium & Medium & Medium & Medium & \includegraphics[width=2cm,height=2cm,keepaspectratio]{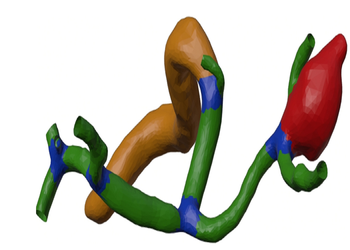} \\
\hline
Voxel~\cite{lei2023cit} & Medium & Low & Low & Easy & \includegraphics[width=2cm,height=2cm,keepaspectratio]{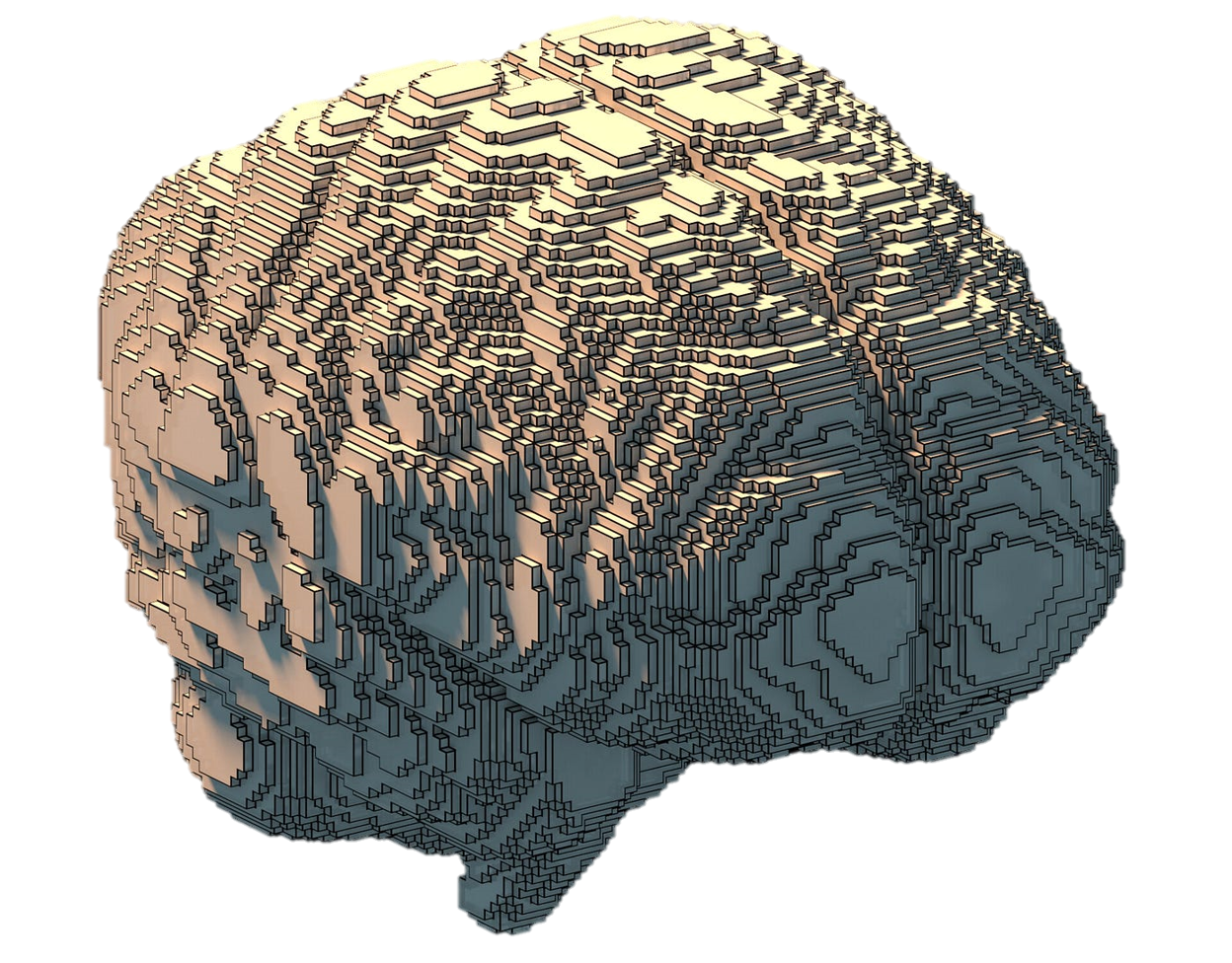} \\
% \hline
% Plenoptic Point Cloud [17, 133] & Huge & High & High & Medium & \includegraphics[width=2cm,height=2cm,keepaspectratio]{example4.png} \\
\hline
Implicit Surfaces~\cite{zha2023endosurf} & Medium & Medium & Medium & Hard & \includegraphics[width=2cm,height=2cm,keepaspectratio]{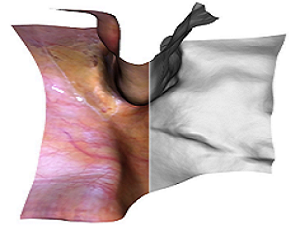} \\
\hline
NeRF~\cite{chen2023cunerf} & Medium & Very high & Very high & Hard & \includegraphics[width=2cm,height=2cm,keepaspectratio]{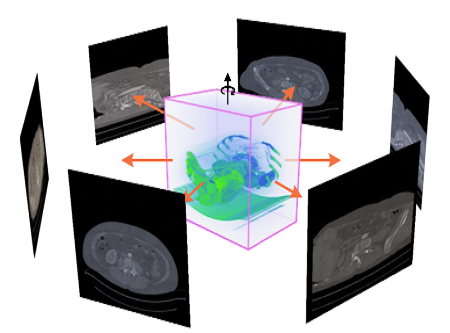} \\
\hline
\end{tabular}
\caption{Comparisons and visualization of different 3D shape representations.}
\label{tab: representation}
\vspace{-10pt}
\end{table*}
% Neural Radiance Fields

\noindent\textbf{Neural Implicit Functions.}
In virtual reality, typical methodologies for 3D shape representations encompass point-based~\cite{yang2020intra}, voxel-based~\cite{lei2023cit}, and mesh-based~\cite{wu2022two} techniques. These approaches explicitly utilize individual points, vertices, or faces to delineate 3D structures, which commonly result in substantial data size. Moreover, they necessitate high-quality raw data for precise reconstruction. A notable limitation of these traditional methods, particularly in medical context, is their inability to produce watertight surfaces, leading to significant drawbacks. 
In contrast, recent advances in neural implicit representation methods have substantially enhanced reconstruction performance.
Neural implicit functions represent a scene by learning a mapping from 3D input coordinates to a shape or surface representation using neural networks, typically through a multi-layer perceptron (MLP) architecture. These functions capture the scene's characteristics in terms of local densities and associated colors, allowing detailed and flexible modeling of complex shapes and surfaces. In~\cite{chen2023cunerf}, an implicit function network was utilized to yield super-resolution medical images at arbitrary scales and free viewpoints in a continuous domain.
For deformable tissues that pose a high requirement for watertight reconstruction, EndoSurf~\cite{zha2023endosurf} was proposed, which effectively learns three neural fields to transform 3D points, predict signed distance functions (SDFs), and estimate colors.

\subsection{Visual Enhancement}
Multiple learning-based methods have been developed to enhance visual perception during the training or surgical phase, improving the skills of surgeons and treatment effect.

\noindent\textbf{Rendering.}
Recent advances in rendering techniques have significantly enhanced rendering performance, leading to improved visual quality of medical data perception. In Semantic Virtual Shadows (SVS)~\cite{sommersperger2023semantic}, instrument-specific shadows are artificially generated, enabling naturally non-existent but important perceptual cues that are present in microscopic surgeries. 
On the other hand, in order to deal with the problem of significant topology changes, occlusion, and spare viewpoints, \textit{neural rendering} has been recently developed to address the limitations of traditional 3D rendering techniques. \textit{Neural rendering} involves using neural networks to capture the complex interactions of scene geometry, lighting, and details, enabling the creation of new views from existing scenes.
In~\cite{wang2022neural}, neural rendering is used to enhance 3D modeling from endoscopic videos, even when there is non-rigid deformation and occlusion. 
% In \cite{corona2022mednerf}, a novel application of neural rendering is presented, enabling the rendering of CT projections from single or multiple-view X-rays.

\noindent\textbf{Perception Enhancement.}
VR devices enable professionals to visualize medical data in a more flexible manner, offering additional cognitive cues during surgical or training procedures. With the aid of AI-powered VR techniques, professionals are able to acquire further information for treatment in virtual environments. 
For instance, a deep learning-based workflow demonstrated in \cite{palumbo2022mixed} supports emergency treatment through automatic segmentation of skin, skull, and ventricles using a 3D U-Net architecture \cite{cciccek20163d}. The segmented data are then integrated into MR images as 3D point clouds, enhancing the precision of medical interpretations. 
Besides, a novel method described in \cite{tamersoy2023accurate} employs deep learning for estimating a patient's height and weight from depth images alone, using an end-to-end single-value regression approach. 
Furthermore, Electrocardio Panorama~\cite{chen2021electrocardio} incorporates vectorcardiography techniques for enhanced visualization, enabling the synthesis of ECG signals from any viewpoint by modeling an underlying electrocardio field.
To provide complementary information for image-guided therapies, \textit{medical image registration} is essential. This process aligns multiple images, volumes, or surfaces within a common coordinate system to identify the common areas.
The study in \cite{ijcai2022p117} introduces a discriminator-free translation network to facilitate the training of the registration network.

\section{VR Medical Data Analytics}
\label{part2}

Enhanced vision in virtual space offers a multitude of visual cues for both human and computer perception, thereby enhancing the analytical capability of medical data. By exploiting the immersive and interactive nature of VR, medical professionals can access a wealth of additional visual information that goes beyond traditional 2D displays. 

\subsection{Structural and Lesion Analysis}
% detection/recognition/localization/segmentation
% interactive 3D visualizations
% 针对重建后的而数据
% In the realm of 3D reconstructed representations, which offer a more comprehensive view than traditional 2D imaging, numerous methodologies have been developed to facilitate enhanced analysis. 
% These methodologies broadly fall into two categories: Structural Analysis and Lesion Analysis. Structural Analysis refers to the granular segmentation and interactive visual analytic examination of systems, organs, tissues, and sub-structures. Besides, Lesion Analysis encompasses the identification, detection, localization, and segmentation of pathological entities such as tumoral lesions, inflammatory lesions, degenerative diseases, and traumatic lesions.
% The reconstructed 3D representations provide a more comprehensive perspective than traditional imaging, particularly for Structural and Lesion Analysis. This encompasses three key technological aspects: detection, segmentation, and interactive visual analytics. 
% 

% In medical scenarios, the use of Virtual Reality (VR) related data, including voxel-based, point-based, and mesh-based formats, offers a more comprehensive perspective compared to traditional 2D imagings, especially in the context of structural and lesion analysis. 
In medical context, the application of VR predominantly utilizes three principal data formats: voxel-based, point-based, and mesh-based. These formats facilitate more sophisticated and nuanced visualization of anatomical structures and disease lesions, which is pivotal for a comprehensive analysis in clinical practice. We provide comparisons and examples of principle data formats in Table~\ref{tab: representation}.

\noindent\textbf{Point-based Format.}
A point cloud employs a large group of individual data points in space to represent 3D objects. Each point contains spatial coordinates and additional attributes (e.g., color in RGB, transparency). This unique characteristic allows straightforward and direct applications of downstream tasks, including object detection and segmentation. However, the use of point clouds also comes with the drawback of high computational resource consumption due to their large data sizes. 
% In Tbale~\ref{tab: representation}, we compare and visualize different 3D shape representations.
In~\cite{liu2023grab,cui2021tsegnet}, operations are directly conducted on point clouds, without requiring further manipulations of the input point cloud data. 
IntrA~\cite{yang2020intra} and 3DTeethSeg~\cite{ben20233dteethseg} are two of the most mainstream point cloud datasets in medical scenarios, and a series of methods in both medical scenarios~\cite{liu2023grab,cui2021tsegnet} and general 3D vision~\cite{wu2019pointconv,li2018pointcnn} were implemented on such datasets, showing the transfer-ability of 3D segmentation methods from the general domain to the medical domain. 
% GRAB-Net\cite{liu2023grab} focus on boosting segmentation performance around anatomical boundaries. 
There are also some works~\cite{zhang2023anatomical} that combine different 3D formats to jointly promote the understanding and analysis of anatomical structures.
In~\cite{zhang2023anatomical}, a point-voxel fusion framework was presented for accurately segmenting the liver into anatomical segments, addressing the challenge of no intensity contrast between adjacent segments by incorporating vessel structure prior.

\noindent\textbf{Mesh-based Format.}
A polygon mesh consists of vertices, edges, and faces that define the shape of a polyhedral object. The 3D mesh format is a collection of meshes that represent the spatial surface, color, and texture of the object. Compared with point clouds, the mesh is suitable for representing complex geometry by a combination of smooth surfaces with a smaller data size. However, the complicated composition of the mesh introduces higher processing costs.
Mesh-based methods~\cite{wu2022two,zhang2021tsgcnet,lian2020deep} take triangle meshes as input and produce corresponding labels for each mesh.
In~\cite{lian2020deep}, cascaded graph-constrained learning modules were introduced for extracting multi-scale local contextual features for automated tooth labeling on raw dental meshes. In~\cite{zhang2021tsgcnet}, a two-stream graph convolutional network was proposed to learn multi-view geometric information from different geometric attributes. 
In~\cite{wu2022two}, a two-stage framework was designed for concurrent segmentation of meshes and regression of anatomical landmarks.
\noindent\textbf{Voxel-based Format.}
Compared with pixels in 2D images, voxels are the most analogous representation in 3D space with a smaller domain gap. Therefore, many 2D methods for medical data processing, by taking into account the additional spatial dimension, can be adapted to encode 3D features and capture the relationships in three-dimensional space. In~\cite{lei2023cit,huang2022scaleformer,dong2022mnet}, voxel-based methods operate on volumetric data represented as a grid of 3D pixels, known as voxels.
% Voxel data adds an additional spatial dimension to the conventional 2D format, resulting in a narrower domain gap. 
CiT-Net~\cite{lei2023cit} was designed as a hybrid architecture of 3D convolutional neural networks (CNNs) hand in hand with 3D Vision Transformers for volumetric medical image segmentation. 
Furthermore, 2D imaging~\cite{salari2023towards}, projections~\cite{wu2022self}, and depth maps~\cite{dima20233d} are frequently employed to facilitate the comprehension and interpretation of voxel-based data. In~\cite{dima20233d}, 2D labels were mapped to the 3D space using depth information to bridge the gap between 3D supervision and 2D supervision for 3D arterial segmentation. In~\cite{salari2023towards}, real-time 2D imaging was leveraged to assist high-precision 3D voxel-based visual landmark detection. 
% \cite{yan20183d} proposes 3D context enhanced region-based CNN (3DCE) to incorporate 3D context information efficiently by aggregating feature maps of 2D projections.

\subsection{Disease Diagnosis}
% classification, 
Recently, based on enhanced visual data, various methods for disease diagnosis have been developed,
which can be grouped into three lines according to their design philosophies.

\noindent\textbf{Extractor-focused Methods.}
This line of methods~\cite{wang2024medtransformer,wu2023d,shao20223d,jang2022m3t} focuses on designing efficient and effective local feature extractors. 
Early work was usually based on using MLPs~\cite{liu20183d}, 3D CNNs~\cite{cciccek20163d}, and GNNs~\cite{zhang2021tsgcnet}.  
% for feature extraction from point-wise data; 3D U-Net~\cite{cciccek20163d}, VNet~\cite{milletari2016v}, and 3D ResNet~\cite{hara2017learning} are mainstream for voxel-wise data. For mesh-wise data, GNNs~\cite{zhou2020graph} are usually used to learn features of nodes and edges, thereby capturing the network's topological structure and local relationships. 
More recently, inspired by the success of Transformers in the general vision domain, Transformer-based backbones~\cite{wang2024medtransformer,wu2023d,shao20223d,jang2022m3t} have gradually unified feature extraction for various 3D medical enhanced visual data.
For example, D-former~\cite{wu2023d} introduced a dilated attention to effectively reduce computation for long-dependent feature extraction process of Transformer.
% 3d volumetric data. 
% \cite{yu20213d} introduces the 3D Medical Point Transformer (3DMedPT), an attention-based model for medical point cloud analysis, which enhances feature representation through position embeddings and Multi-Graph Reasoning.

\noindent\textbf{Semantics-enhanced Methods.}
Semantics-based methods~\cite{pan2023combat,zhao2023multi,xia2019novel} are concerned with integrating various intrinsic information sources to enhance the comprehension of the underlying semantics associated with features and target objects. 
This can involve leveraging  
% patient medical histories, genetic profiles, or population health data 
multi-view consistency~\cite{pan2023combat},
multi-task consistency~\cite{zhao2023multi},
and multi-modality consistency~\cite{xia2019novel}
to improve the accuracy of diagnostic algorithms. 
% \cite{zhao2023multi} proposes to jointly segment the tumor and classifies abnormality in a multi-task manner. 
In~\cite{pan2023combat}, a multi-view relation-aware consistency and virtual feature compensation (MRC-VFC) framework was proposed for long-tailed medical image classification.

\noindent\textbf{Knowledge-embedded Methods.}
Knowledge-embedded methods~\cite{rygiel2023centerlinepointnet,wu2023gcl} aim to incorporate external knowledge and priors to enable networks to conduct diagnosis in a more informative and effective manner. 
% By integrating databases of clinical outcomes, pathology reports, and expert annotations into the learning process, these methods can align the diagnostic model's predictions with established medical understanding. 
% Consequently, such knowledge-aided approaches can bridge the gap between pure data-driven models and the nuanced decision-making used in human expert diagnostics, facilitating advanced AI assistance in patient care.
% Based on 3D cardiac shape prior, ~\cite{beetz2023multi} presents a multi-objective point cloud autoencoder as a novel geometric deep learning approach for explainable infarction prediction. 
Based on geometric priors, a geometry-based architecture was proposed to utilize implicit geometry embedding for directly estimating hemodynamic features~\cite{rygiel2023centerlinepointnet}.
Meta labels~\cite{wu2023gcl}, representing specific attribute information of medical images, serve as valuable cues for networks to enhance the identification and classification of pathological findings.
Such incorporation of domain-specific knowledge not only enhances the interpretability of AI-driven diagnostics but also increases the reliability of systems in clinical settings. 
% geometry prior

% meta label 

\subsection{Surgical Support}
An important application scenario of AI-powered VR is in surgical support, where it can offer surgeons more precise planning, real-time tracking, and reliable feedback throughout the surgical procedures.

\noindent \textbf{Pre-operative Planning.}
Pre-operative planning is a crucial phase in surgery, involving detailed patient-specific analysis and strategic procedure mapping to ensure optimal surgical outcomes and minimize risks.  
VR-enhanced planning in surgery primarily serves two key purposes: (1) facilitating the arrangement of surgical procedures~\cite{bukas2021patient}, and (2) enabling the simulation of surgical outcomes~\cite{friedrich2023point}.
% \cite{kordon2021automatic} investigates to assist the trauma surgeon in tibial guide pin placement and tunnel drilling in the technically demanding PCL reconstruction surgery.
In~\cite{bukas2021patient}, a personalized automatic high-level framework was presented for planning osteoplasty procedures.
In~\cite{friedrich2023point}, a novel approach was proposed for implant generation based on a combination of 3D point cloud diffusion models and voxelization networks.
% \cite{chen2023semantic} propose a semantic-guided and knowledge-based generative framework to predict the visual outcome of orthodontic treatment. 

% \cite{fang2023soft} proposes a soft-tissue-driven framework that can automatically create and verify surgical plans.

\noindent \textbf{Real-time Object Recognition and Tracking.}
% In the surgical context, based on the different types of targets, real-time object recognition and tracking can be broadly categorized into two classes: lesion-centric and device-centric.
% Lesion-centric recognition and tracking, accurately measuring monitoring pathological changes intra-operatively. 
% Conversely, device-centric one focuses on real-time localization and control of surgical instruments, enhancing precision and safety in procedures. 
% Integrating technologies like RFID\cite{} and optical systems\cite{}, this technique is transformative in minimally invasive surgeries. 
Real-time object recognition and tracking play a crucial role in surgical procedures, ensuring immediate and accurate identification of surgical instruments and anatomical structures.
Based on the categories of the targets, existing works can be broadly classified into two types: lesion-centric and device-centric.
% In the surgical setting, real-time object recognition and tracking 
Lesion-centric works~\cite{szeskin2023liver,cai2021deep} involve accurate measurement and monitoring of pathological changes during surgical procedures.  
In~\cite{szeskin2023liver}, a fully automatic end-to-end pipeline was presented for liver lesion change analysis in consecutive (prior and current) abdominal CECT scans of oncology patients.
Highlighting unusual lesion labels and lesion change patterns helps radiologists identify possibly overlooked or faintly visible lesions.
In contrast, device-centric approaches~\cite{gsaxner2021inside} focus on real-time localization and control of surgical instruments.
% This technique integrates technologies such as RFID and optical systems, making it transformative in the context of minimally invasive surgeries.
% \cite{budd2023deep} proposes an autofocusing method based on deep reinforcement learning to mitigate the issue of reduced focal depth in real-time HSI systems. 
In~\cite{gsaxner2021inside}, 6-DOF tracking was novelly achieved by purely utilizing on-board stereo cameras of a HoloLens2 to track the same retro-reflective marker spheres used by current optical navigation systems.

\noindent \textbf{Post-operative Feedback.}
Post-operative feedback is essential in evaluating the success of surgical procedures, guiding patient recovery, and informing future improvements in surgical practice and patient care, where motion analysis and outcome assessment are two key technical aspects. 
Motion analysis~\cite{nwoye2022rendezvous,twinanda2016endonet} refers to precisely detecting fine-grained actions of surgeons, with a particular focus on their hand movement.
Rendezvous (RDV)~\cite{nwoye2022rendezvous} was proposed to recognize surgical action triplets (i.e., $<$instrument, verb, target$>$) in endoscopic videos by leveraging attention at the spatial and semantic levels, respectively.
POV-Surgery dataset~\cite{wang2023pov} is a large-scale, synthetic, egocentric dataset focusing on pose estimation of hands with different surgical gloves and orthopedic surgical instruments.
Outcome assessment~\cite{jamzad2023bridging} involves systematic evaluation of the results and impact of a treatment or intervention, measuring its effectiveness, safety, and patient satisfaction.

% This continuous monitoring of movements facilitates real-time localization within the surgical process, enabling the system to provide timely assessments and feedback. This approach is pivotal in enhancing the precision and efficacy of surgical procedures, offering a critical tool for quality assurance and skill development in the operating theater.
% In this process, real-time responsiveness is a critical factor, as it enables surgeons to make timely adjustments in their actions for the next step of the procedure. This immediacy is vital for ensuring the precision and safety of surgical interventions, allowing for instantaneous adaptation to the evolving demands of the operation.

% Treatment Planning
% Disease Prognosis
% Clinical Decision Support
% Patient Care
% skill assessment
% Simulation-Based Learning

\section{VR-assisted Intervention}
\label{part3}
Based on combination of AI-powered visualization and analysis technologies, VR-assisted intervention technologies have significantly enhanced professionals' capabilities during the inspection and surgical phases. This section explores two key applications of VR-assisted intervention: \textit{Intra-operative Intervention} and \textit{Interactive Collaboration}.

%Local & Remote intervention
%visual-based or ??
\subsection{Intra-operative Intervention}
Intra-operative intervention focuses on leveraging VR technologies to enhance professionals' abilities during surgery. VR platforms attached with AI algorithms offer additional guidance/assistance during the complex intervention phases. Leveraging the advantages of augmented vision, a virtual assistance platform based on commodity optical see-through head-mounted displays (OST-HMDs) was implemented~\cite{doughty2021surgeonassist}, which provides users with action-and-workflow-driven assistance with near-real-time performance on the Microsoft HoloLens2 OST-HMD. 
% To achieve such performance, an efficient convolutional neural network (CNN) backbone is exploited to extract discriminative features from image data and a low-parameter recurrent neural network (RNN) architecture is used to learn long-term temporal dependencies. 
To achieve real-time performance, CNN for spatial feature extraction and RNN for temporal relationship learning were introduced to form the core of the system's analytical capabilities. 
A similar attempt~\cite{palumbo2022mixed} was also implemented on the same platform (Microsoft HoloLens2) in the emergency treatment scenario, where 3D U-Net~\cite{cciccek20163d} architecture is exploited for skill, face skin, and ventricle segmentation. These works provide valuable insights into developing real-time surgical guidance systems based on VR platforms. 
Furthermore, regarding surgical scene understanding and reasoning, which serve as a backbone of VR-assisted intervention, phase recognition~\cite{ding2022exploring} and scene captioning~\cite{beddiar2023automatic} techniques play a crucial role, aiding surgeons to obtain direct responses and comprehensive understanding of the current state of surgery. This task encompasses real-time processing and integration of diverse data streams, including visual inputs from the surgical field, patient vitals, and historical medical data.
%继续写recoginition 和 scene captioning

\subsection{Interactive Collaboration}
Recent advances in human-machine collaboration have significantly reduced human effort in field operations, thereby fostering widespread implementation of AI-aided interactive collaboration. At the forefront of these advances is Interactive Query. In~\cite{seenivasan2023surgicalgpt}, a Vision Question Answering (VQA) system was proposed specifically for medical scenarios, providing a robust and reliable surgical visual question answering solution that can respond to questions by inferring from context-enriched surgical scenes. Beyond that, in~\cite{bai2023surgical}, Visual Question Localized-Answering (VQLA) was further developed that offers more guidance by highlighting the specific areas in the images related to the question and answer. This allows a better
understanding of complex medical diagnoses and surgical scenes. Notably, Scene Graph~\cite{holm2023dynamic,ozsoy2023holistic} is a more holistic, semantically meaningful, and human-readable way to represent surgical videos while encoding all anatomical structures, tools, and their interactions. 

Datasets play a critical role in this domain. For instance, in~\cite{sharghi2020automatic}, a dataset was created to capture different robot-assisted interventions, focusing on the phase recognition of the OR scene. The MVOR dataset~\cite{srivastav2018mvor} contains synchronized multi-view frames from real surgeries with human pose annotations. This dataset facilitates research on human motion analysis and understanding during surgical procedures. Additionally, the 4D-OR dataset~\cite{ozsoy20224d} is the first publicly available 4D surgical semantic scene graphs dataset, contributing to advancements in surgical planning and decision-making.

\section{Challenges and Future Prospects}
\subsection{Challenges}

\noindent \textbf{Data and Integration Limitations.}
A primary challenge in AI-driven VR for medical applications lies in the quality and availability of data. High-quality, diverse datasets are essential for training effective AI models. Yet, such data currently are still scarce and fragmented. Additionally, integrating the advanced AI-VR technologies with existing healthcare systems and workflows is a complex task, requiring both technical compatibility and operational adjustments.

\noindent \textbf{Ethical and Legal Issues.} 
Ethical issues, particularly concerning with patient privacy, data security, and informed consent, are crucial. AI-VR systems must comply with stringent healthcare regulations to ensure patient confidentiality and safety. Furthermore, the issue of liability in the cases of AI-VR related errors remains unresolved, complicating the legal landscape for healthcare providers, technology developers, and governments.

\noindent \textbf{Technological and User Acceptance.} 
Achieving realism and accuracy in VR simulations is vital for effective medical training and treatment, yet it remains technically challenging. The `black box' nature of some AI systems can impede user trust and acceptance, as medical professionals often need to understand the decision-making processes behind AI recommendations. Additionally, designing user-friendly interfaces for a diverse range of users and overcoming technophobia among healthcare professionals and patients is critical for the widespread adoption of AI-driven VR systems.

\subsection{Future Prospects}

\noindent \textbf{Immersive Medicine through Technological Advancements.} 
Future advancements in AI algorithms enable VR technologies to provide more accurate, efficient solutions in immersive medical services. For example, the integration of Natural Language Processing (NLP) approaches within VR environments will facilitate an enhanced understanding of patient speech and text records. Besides, advanced interactive sonification technologies~\cite{matinfar2023tissue} can also lead to more comprehensive and detailed diagnostics, thereby enhancing the accuracy and effectiveness of diagnoses and treatments.
% This capability is particularly pivotal in psychological diagnostics and therapeutic interventions.

% This progress will facilitate personalized medicine, where AI-driven VR experiences are tailored to individual patient needs, enhancing education, therapy, and treatment planning.
\noindent \textbf{Tailored Therapeutic Interventions with AI-Driven Analytics.}
Utilization of AI algorithms in analyzing patient interactions within VR scenarios enables the customization of therapeutic approaches and options, especially in mental healthcare settings. Furthermore, the integration of biofeedback into VR environments, where artificial intelligence (AI) adapts experiences according to real-time physiological data, signifies a remarkable advancement in tailoring patient care and treatment experience.

\noindent \textbf{AI-Enhanced Real-Time Clinical Analytics.} 
% AI-powered VR exhibits the potential to revolutionize tele-medicine and remote medical training, particularly in under-served or under-developed areas. Incorporating real-time analytical capabilities within VR, AI algorithms will provide instantaneous diagnostic support, elevating the decision-making processes in clinical and non-clinical settings. Besides, by providing high-quality medical services remotely, these technologies hold the potential to bridge the gaps in healthcare accessibility, quality, and equality.
AI-powered VR has the potential to transform telemedicine and remote medical training, especially in underserved areas. By integrating real-time analytics, AI algorithms can offer immediate diagnostic support, improving decision-making in both clinical and non-clinical settings. Additionally, these technologies can enhance healthcare accessibility, quality, and equity by delivering high-quality medical services remotely.

\section*{Acknowledgments}
This research was partially supported by National Natural Science Foundation of China under grants No. 62176231, No. 62106218, No. 82202984, No. 92259202 and No. 62132017, Zhejiang Key R\&D Program of China under grant No. 2023C03053.
\subsubsection{Contribution Statement}
Yixuan Wu and Kaiyuan Hu contributed equally to this work. Jian Wu and Danny Z. Chen are the corresponding authors.
% , responsible for the overall supervision and coordination of the project.

% \noindent \textbf{Interdisciplinary Collaboration and Regulatory Evolution. } The effective development and deployment of AI-VR solutions in healthcare will require close collaboration between technologists, healthcare professionals, and policymakers. Ethical AI development, focusing on transparency and responsibility, must be a priority. Additionally, the evolution of regulatory frameworks is essential to balance innovation with patient safety and privacy.

% \noindent \textbf{Long-Term Impact and Research. } Continued research into the long-term efficacy, safety, and socioeconomic impacts of AI-VR applications in medicine will be crucial. This includes studying their effects on healthcare costs, access, quality, and the overall patient and practitioner experience.
%% The file named.bst is a bibliography style file for BibTeX 0.99c
\bibliographystyle{named}
\bibliography{ijcai24}

\end{document}